\documentclass[10pt,letterpaper,twocolumn]{article}
\PassOptionsToPackage{hyphens}{url}
\usepackage[latin9]{inputenc}
\usepackage{array}
\usepackage{textcomp}
\usepackage{url}
\usepackage{multirow}
\usepackage{amsmath}
\usepackage{amssymb}
\usepackage{graphicx}
\usepackage[unicode=true,
 bookmarks=false,
 breaklinks=false,pdfborder={0 0 1},backref=section,colorlinks=false]
 {hyperref}
\hypersetup{
 pagebackref,breaklinks,colorlinks}

\makeatletter

\pdfpageheight\paperheight
\pdfpagewidth\paperwidth

\DeclareTextSymbolDefault{\textquotedbl}{T1}
\providecommand{\tabularnewline}{\\}

\usepackage[pagenumbers]{cvpr} 

\usepackage{graphicx}

\usepackage{import}

\usepackage[capitalize]{cleveref}
\crefname{section}{Sec.}{Secs.}
\Crefname{section}{Section}{Sections}
\Crefname{table}{Table}{Tables}
\crefname{table}{Tab.}{Tabs.}

\def\cvprPaperID{****} 
\def\confName{CVPR}
\def\confYear{2023}

\makeatother

\begin{document}

\title{Rawgment: Noise-Accounted RAW Augmentation Enables Recognition in
a Wide Variety of Environments}

\author{
Masakazu Yoshimura\quad Junji Otsuka\quad Atsushi Irie\quad Takeshi Ohashi\\
Sony Group Corporation\\
{\tt\small \{masakazu.yoshimura, junji.otsuka, atsushi.irie, takeshi.a.ohashi\}@sony.com}%
}

\maketitle
\begin{abstract}
Image recognition models that work in challenging environments (e.g.,
extremely dark, blurry, or high dynamic range conditions) must be
useful. However, creating training datasets for such environments
is expensive and hard due to the difficulties of data collection and
annotation. It is desirable if we could get a robust model without
the need for hard-to-obtain datasets. One simple approach is to apply
data augmentation such as color jitter and blur to standard RGB (sRGB)
images in simple scenes. Unfortunately, this approach struggles to
yield realistic images in terms of pixel intensity and noise distribution
due to not considering the non-linearity of Image Signal Processors
(ISPs) and noise characteristics of image sensors. Instead, we propose
a noise-accounted RAW image augmentation method. In essence, color
jitter and blur augmentation are applied to a RAW image before applying
non-linear ISP, resulting in realistic intensity. Furthermore, we
introduce a noise amount alignment method that calibrates the domain
gap in the noise property caused by the augmentation. We show that
our proposed noise-accounted RAW augmentation method doubles the image
recognition accuracy in challenging environments only with simple
training data.
\end{abstract}

\section{Introduction}

\label{sec:intro}

Although image recognition has been actively studied, its performance
in challenging environments still needs improvement \cite{Hong2021Crafting}.
Sensitive applications such as mobility sensing and head-mounted wearables
need to be robust to various kinds of difficulties, including low
light, high dynamic range (HDR) illuminance, motion blur, and camera
shake. One possible solution is to use image enhancement and restoration
methods. A lot of DNN-based low-light image enhancement \cite{lv2018mbllen,zhang2019kindling,jiang2021enlightengan,wei2018deep,guo2020zero,ma2022toward},
denoising \cite{zhang2022idr,monakhova2022dancing,tu2022maxim}, and
deblurring \cite{zhang2022pixel,whang2022deblurring,tu2022maxim}
methods are proposed to improve the pre-captured sRGB image quality.
While they are useful for improving pre-captured image quality, a
recent work \cite{Hong2021Crafting} shows that using them as preprocessing
for image recognition models has limited accuracy gains since they
already lost some information, and restoring the lost information
is difficult.

\begin{figure}
\centering

\def\svgwidth{1.0\columnwidth}

\scriptsize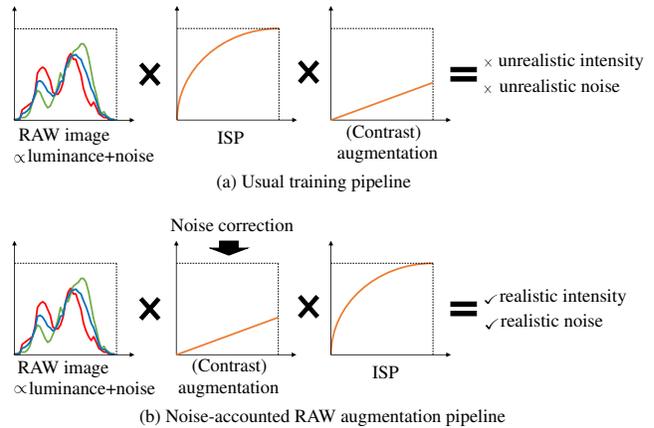

\caption{The concept of the proposed noise-accounted RAW augmentation. Conventional
augmentation (a) is applied to the output of an ISP; due to the nonlinear
operations in the ISP, it produces images that cannot be captured
at any ambient light intensities. Instead, ours (b) applies augmentation
before an ISP. It generates realistic pixel intensity distribution
that can be captured when the light intensity is different. Moreover,
the noise amount is also corrected to minimize the domain gap between
real and augmented ones.}

\label{fig:rawaug}
\end{figure}

Another possible solution is to prepare a dataset for difficult environments
\cite{morawski2022genisp,chen2018learning}. However, these datasets
only cover one or a few difficulties, and creating datasets in various
environments is too expensive. Especially, manual annotation of challenging
scenes is difficult and time-consuming. For example, we can see almost
nothing in usual sRGB images under extremely low-light environments
due to heavy noise. In addition, some regions in HDR scenes suffer
from halation or blocked-up shadows because the 8-bit range of usual
sRGB images cannot fully preserve the real world, which is 0.000001
$[cd/m^{2}]$ under starlight and 1.6 billion $[cd/m^{2}]$ under
direct sunlight \cite{reinhard2010high}. Heavy motion blur and camera
shake also make annotation difficult. Some works capture paired short-exposure
and long-exposure images, and the clean long-exposure images are used
for annotation or ground truth \cite{Hong2021Crafting,ignatov2017dslr,ignatov2020replacing,jiang2019learning}.
The limitation is that the target scene needs to be motionless if
the pairs are taken sequentially with one camera \cite{Hong2021Crafting},
and positional calibration is required if the pairs are taken with
synchronized cameras \cite{ignatov2017dslr,ignatov2020replacing}.
Some works use a beam splitter to capture challenging images and their
references without calibration \cite{wang2022neural,jiang2019learning}.
However, they are difficult to apply in dark scenes. Moreover, HDR
images cannot be taken in the same way because some regions become
overexposed or underexposed in both cameras.

To this end, we aim to train image recognition models that work in
various environments only using a training dataset in simple environments
like bright, low dynamic range, and blurless. In this case, image
augmentation or domain adaptation is important to overcome the domain
gap between easy training data and difficult test data. However, we
believe usual augmentations on sRGB space are ineffective because
it does not take into account the nonlinear mapping of ISPs. In particular,
tone mapping significantly changes the RAW image values, which are
roughly proportional to physical brightness \cite{wang2020practical}.
Contrast, brightness, and hue augmentation on sRGB space result in
unrealistic images that cannot be captured under any ambient light
intensity as shown in Fig. \ref{fig:rawaug}(a). In contrast, we propose
augmentation on RAW images. In other words, augmentation is applied
before ISPs to diminish the domain shift as shown in Fig. \ref{fig:rawaug}(b). 

Oher possible sources of the domain gap are differences in noise amount
and noise distribution. To tackle these problems, we propose a method
to align both light intensity and noise domains. Recent works show
that adding physics-based realistic noise improves the performance
of DNN-based denoisers \cite{wang2020practical,wei2020physics,brooks2019unprocessing,zamir2020cycleisp}
and dark image recognition \cite{Hong2021Crafting,cui2021multitask}.
Although their proposed sensor noise modelings are accurate, they
assume that the original bright images are noise free. In contrast,
we propose to modify the noise amount after contrast, brightness,
and hue conversion considering the noise amount in the original images.
It enables a more accurate alignment of the noise domain. Even bright
images may have dark areas due to shadows or object colors, and their
prior noise cannot be ignored. Another merit of our method is that
it is possible to take dark images that already contain a lot of noise
as input. In addition to noise alignment after color jitter augmentation,
we show the importance of noise alignment after blur augmentation,
which is proposed for the first time in this paper.

Our contributions are as follows:
\begin{itemize}
\item It is the first work to emphasize the importance of augmentation before
ISP for image recognition to the best of our knowledge.
\item Noise amount alignment method is proposed to reduce the noise domain
gap after RAW image augmentation. In contrast to previous works, our
proposed method takes into account prior noise in the input image.
It enables more accurate alignment and use of any strength of augmentation
and even already noisy input.
\item We show qualitative analysis for the validity of our sensor noise
modeling and corresponding noise-accounted augmentation. We prove
that our proposed noise-accounted RAW augmentation has the edge over
the previous methods.
\end{itemize}

\section{Related Works}

\subsection{Recognition in Difficult Environment}

Many works have tackled image recognition in difficult environments.
For low-light environments, several works improve the accuracy by
replacing a traditional ISP with a powerful DNN-based ISP to create
clean images for downstream image recognition models \cite{diamond2021dirty,morawski2022genisp,liu2022deep}.
Even though these methods are promising because there is no information
loss, the computational cost is the problem. Another approach is direct
RAW image recognition without ISP \cite{schwartz2021isp,Hong2021Crafting}.
Their image recognition models benefit from the richest information
and improve accuracy under low-light environments. However, several
works report ISPs, especially tone mapping, are helpful for machine
vision \cite{wu2019visionisp,hansen2021isp4ml}. Direct RAW image
recognition may works well if the images have a low dynamic range.
Another approach is domain adaptation or related methods which support
low-light recognition with bright images \cite{sasagawa2020yolo,Hong2021Crafting,cui2021multitask}.

For HDR environments, some works propose DNN-based auto-exposure control
\cite{tomasi2021learned,onzon2021neural} to improve downstream recognition.
Also, multi-frame HDR synthesis methods \cite{dudhane2022burst,bhat2021deep}
can be used as a preprocessing, although camera motion makes them
challenging. A luminance normalization method is also introduced to
improve recognition performance under varying illumination conditions
\cite{jenicek2019no}.

For blurry environments, deblurring methods are actively studied \cite{whang2022deblurring,zhang2019deep}.
These DNN-based methods successfully restore clear images from heavily
blurred images.

Differing from the above, we aim to do image recognition under all
the above difficulties using simple scene training data with our proposed
augmentation method. We do not use domain adaptation methods since
these methods are usually used in a setting where the target domain
is equal to or smaller than the source domain \cite{goodfellow2014generative}.
On the contrary, in our setting, the distribution of the target domain
is much wider than that of the source domain.

\subsection{Image Conversion on RAW}

Recently, several methods \cite{brooks2019unprocessing,lv2018mbllen,cui2021multitask}
convert bright sRGB images into realistic dark images by the following
procedures. First, they invert an ISP pipeline to generate RAW-like
images followed by illumination change on the RAW data space with
plausible sensor noise. Afterward, degraded sRGB is generated by applying
the forward ISP pipeline. This operation avoids the nonlinear mappings
in the ISP and simulates short exposures and dark environments. With
a similar intention, we propose to apply augmentation before ISPs
to train image recognition models. 

\subsection{Noise Modeling and Noise Amount Alignment}

In the electronic imaging sensor community, detailed noise modelings
based on electric current and circuit have been studied \cite{suh2010column,el1998modeling,konnik2014high,gow2007comprehensive}.
They are precise but difficult to be applied to the image-to-image
conversion. Thus, in the machine vision community, simplified pixel
value-based noise modelings are proposed based on electric noise modelings
\cite{wang2020practical,wei2020physics,brooks2019unprocessing}. Although
the noise model of \cite{wei2020physics} is well designed with a
high degree of freedom Tukey lambda distribution \cite{joiner1971some},
we are based on the well-established heteroscedastic Gaussian model
\cite{brooks2019unprocessing,punnappurath2022day,foi2008practical,zamir2020cycleisp}
because it is still well fitted to real sensor noise and we can consider
prior noise in original images which will be explained later.

Recently, adding realistic model-based sensor noise to the ground
truth clean images is proved to be helpful to train DNN-based denoiser
\cite{wang2020practical,zamir2020cycleisp,wei2020physics,brooks2019unprocessing}
and low-light object detection models \cite{Hong2021Crafting,cui2021multitask}.
Although they use highly consistent noise models, they regard original
images as noise-free. In contrast, we propose to modify the noise
amount after image conversion considering the noise amount of the
original images. It enables a more accurate alignment of the noise
domain and enables the use of any intensity of augmentation and already
noisy images as input.

\section{Methodology}

In this section, we introduce our noise model, calibration procedure,
and proposed noise-accounted RAW augmentation.

\subsection{Noise Model}

First, we briefly introduce our noise model for later explanation
although it is based on the well-established heteroscedastic Gaussian
model \cite{brooks2019unprocessing,punnappurath2022day,foi2008practical,zamir2020cycleisp}.
The number of photons $u$ hitting the photodiode of each pixel is
converted to a voltage with quantum efficiency $\alpha$. Some processing
is then performed to read out the voltage, in which noise $n_{d}$
is inevitably mixed. Next, analog gain $g$ is multiplied to amplify
the value. Lastly, the voltage is converted to a digital value. We
simplify and summarize the noise after analog gain as $n_{r}$. Since
it is common to use analog gain which has a better signal-to-noise
ratio (SNR), we omit the digital gain term in our noise model. To
sum up, the photon-to-RAW pixel value conversion can be formulated
as, 
\begin{equation}
x=g\left(\alpha u+n_{d}\right)+n_{r}.\label{eq:1}
\end{equation}

We approximate $n_{d}$ and $n_{r}$ as Gaussian noise $\mathcal{N}\left(0,\sigma_{d}^{2}\right)$
$\mathcal{N}\left(0,\sigma_{r}^{2}\right)$ and the number of photons
$u$ itself obeys the Poisson distribution $\mathcal{P}\left(\bar{u}\right)$
where $\bar{u}$ is the expected number of photons. If $\bar{u}$
is large enough, we can approximate as $\mathcal{P}\left(\bar{u}\right)\fallingdotseq\mathcal{N}\left(\bar{u},\bar{u}\right)$
\cite{foi2008practical}. Thus, our noise model is as follows; 
\begin{equation}
x\sim g\left(\alpha\mathcal{N}\left(\bar{u},\bar{u}\right)+\mathcal{N}\left(0,\sigma_{d}^{2}\right)\right)+\mathcal{N}\left(0,\sigma_{r}^{2}\right).\label{eq:2}
\end{equation}
 We show the validity of the Gaussian approximation of $n_{d}$, $n_{r}$,
and $\mathcal{P}\left(\bar{u}\right)$ in the Section \ref{subsec:Calibration-of-the}.
We don't follow the further development of the formula in \cite{foi2008practical}
for our purpose.

Gaussian distribution has the following convenient natures;

\begin{equation}
\begin{cases}
X\sim\mathcal{N}\left(\mu_{X},\sigma_{X}^{2}\right)\\
Y\sim\mathcal{N}\left(\mu_{Y},\sigma_{Y}^{2}\right)\\
X+Y\sim\mathcal{N}\left(\mu_{X}+\mu_{Y},\sigma_{X}^{2}+\sigma_{Y}^{2}\right)\\
cX\sim\mathcal{N}\left(c\mu_{X},c^{2}\sigma_{X}^{2}\right)
\end{cases},\label{eq:gauss}
\end{equation}
if $X$ and $Y$ are independent, and that's why we choose the simple
Gaussian approximation instead of the recently proposed more expressive
noise model \cite{wei2020physics}. These natures enable the proposed
noise-accounted RAW augmentation to account for prior noise in input
images. Furthermore, they further simplify our noise model as, 
\begin{equation}
x\sim\mathcal{N}\left(g\alpha\bar{u},g^{2}\alpha^{2}\bar{u}+g^{2}\sigma_{d}^{2}+\sigma_{r}^{2}\right).\label{eq:4}
\end{equation}
 Because the expected number of photon $\bar{u}$ is inconvenient
to use in image-to-image conversion, we replace it with the expected
pixel value $\mu_{x}=g\alpha\bar{u}$, and our final noise model is
defined as, 
\begin{equation}
x\sim\mathcal{N}\left(\mu_{x},\sigma_{x}^{2}=g\alpha\mu_{x}+g^{2}\sigma_{d}^{2}+\sigma_{r}^{2}\right).\label{eq:noise_model}
\end{equation}

\subsection{Noise Model Calibration }

\begin{figure*}
\centering

\def\svgwidth{1.85\columnwidth}

\scriptsize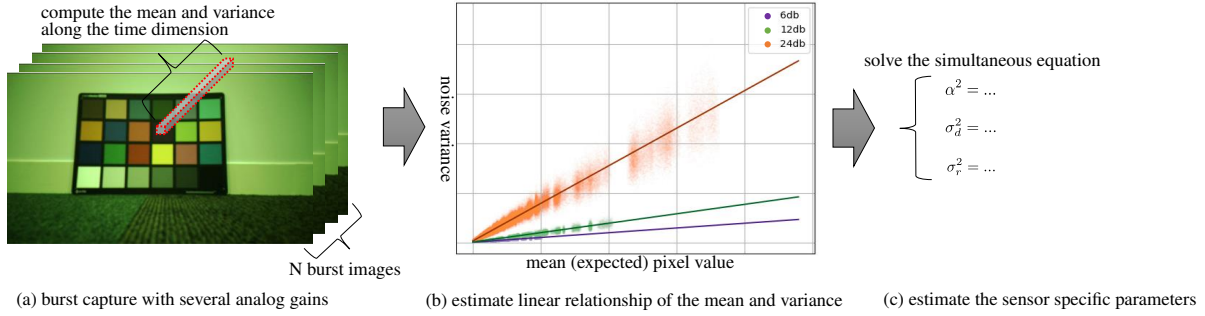

\caption{Our noise model calibration procedures to a target sensor. It just
needs to capture burst RAW images and does not need special devices.}

\label{fig:calibration}
\end{figure*}

Our sensor noise model shown in Eq. (\ref{eq:noise_model}) has three
parameters, $\alpha$, $\sigma_{d}^{2}$, and $\sigma_{r}^{2}$, which
have to be calibrated per target sensor. We capture a series of RAW
images of a color checker as shown in Fig. \ref{fig:calibration}(a).
We then calculate the mean $\mu_{x}$ and variance $\sigma_{x}^{2}$
along the time direction of each pixel position. We calculate them
along the time direction instead of the spatial direction as performed
in \cite{wang2020practical} since lens distortion changes the luminance
of the same color patch. These operations are performed several times
by changing the analog gain and exposure time. Eventually, we get
various sets of $\left\{ \mu_{x},\sigma_{x}^{2}\right\} $ for each
analog gain. Note that we calculate mean and variance without separating
RGB channels because there is no significant difference in noise properties.

In Eq. (\ref{eq:noise_model}), $\mu_{x}$ and $\sigma_{x}^{2}$ have
a linear relationship per analog gain $g_{n}$; 
\begin{equation}
\sigma_{x}^{2}=a_{g_{n}}\mu_{x}+b_{g_{n}}.\label{eq:6}
\end{equation}
Therefore, we solve linear regression to estimate $a_{g_{n}}$ and
$b_{g_{n}}$ per gain like Fig. \ref{fig:calibration}(b). In addition,
we use RANSAC \cite{fischler1981random} to robustly take care of
outlier $\left\{ \mu_{x},\sigma_{x}^{2}\right\} $ pairs.

Finally, we estimate $\alpha$, $\sigma_{d}^{2}$, and $\sigma_{r}^{2}$
from the following redundant simultaneous equations by least-squares
method,

\begin{equation}
\begin{cases}
a_{g_{1}}=g_{1}\alpha\\
\:\:\:\vdots\\
a_{g_{n}}=g_{n}\alpha
\end{cases},\label{eq:7-1}
\end{equation}

\begin{equation}
\begin{cases}
b_{g_{1}}=g_{1}^{2}\sigma_{d}^{2}+\sigma_{r}^{2}\\
\:\:\:\vdots\\
b_{g_{n}}=g_{n}^{2}\sigma_{d}^{2}+\sigma_{r}^{2}
\end{cases}.\label{eq:8}
\end{equation}
The procedure above enables calibration of the sensor noise model
without special devices. We later show that our sensor model and the
calibration method represent the real sensor noise with enough preciseness.

\subsection{Noise-Accounted RAW Augmentation}

We propose augmentation before ISP instead of the usual augmentation
after ISP to generate realistic images. Furthermore, we improve the
reality of the augmented images by considering the sensor noise model.
Unlike the previous works \cite{wang2020practical,wei2020physics,cui2021multitask,Hong2021Crafting,brooks2019unprocessing,zamir2020cycleisp},
ours takes the prior noise amount of input images into account. It
generates more realistic noise since even bright images have some
extent of noise. Especially, dark parts due to shadow or the color
of objects might have a non-negligible amount of noise. Moreover,
it allows any brightness of input images different from previous works.
Specifically, we introduce how to adjust noise amount after contrast,
brightness, hue, and blur augmentation.

\subsubsection{Color Jitter Augmentation}

Contrast, brightness, and hue augmentation simulate different exposure
times, light intensities, and analog gain. Hence, we first assume
to multiply the exposure time, light intensity, and analog gain by
$p_{e}$, $p_{i}$, and $p_{g}$ respectively. Because $p_{e}$ and
$p_{i}$ equally change the number of photon $u$ in the case of our
noise model, we rewrite them as $p_{u}=p_{e}p_{i}$. Then, images
in the above environment settings $x_{new}$ can be rewritten as,
\begin{equation}
x_{new}\sim\mathcal{N}\left(\begin{split}(p_{g}g)\alpha(p_{u}\bar{u}),\\
(p_{g}g)^{2}\alpha^{2}(p_{u}\bar{u})+(p_{g}g)^{2}\sigma_{d}^{2}+\sigma_{r}^{2}
\end{split}
\right).\label{eq:9}
\end{equation}
 Based on Eq. (\ref{eq:gauss}), it can be expanded as 
\begin{align}
x_{new} & \sim\mathcal{N}\left((p_{g}p_{u})\alpha g\bar{u},\:(p_{g}p_{u})^{2}(g^{2}\alpha^{2}\bar{u}+g^{2}\sigma_{d}^{2}+\sigma_{r}^{2})\right)\nonumber \\
 & \:\:\:\;+\mathcal{N}\left(\right.0,\:-(p_{g}p_{u})^{2}(g^{2}\alpha^{2}\bar{u}+g^{2}\sigma_{d}^{2}+\sigma_{r}^{2})\label{eq:10}\\
 & \:\:\:\:\;\:\:\;\:\:\:\:\:\:\;\:\:\:\:\:\:\:+(p_{g}g)^{2}\alpha^{2}(p_{u}\bar{u})+(p_{g}g)^{2}\sigma_{d}^{2}+\sigma_{r}^{2}\left.\right).\nonumber 
\end{align}
 By inserting $\mu_{x}=g\alpha\bar{u}$ and original pixel value,
$x_{pre}\sim\mathcal{N}\left(\mu_{x},g\alpha\mu_{x}+g^{2}\sigma_{d}^{2}+\sigma_{r}^{2}\right)$,
it can be expressed with a pixel value-based equation as follows;
\begin{align}
x_{new} & \sim p_{u}p_{g}x_{pre}+\nonumber \\
 & \:\:\:\;\mathcal{N}(0,\:p_{u}(1-p_{u})p_{g}^{2}g\alpha\mu_{x}\label{eq:11}\\
 & \:\;\:\:\;\:\:\:\:\:\:\;\:\:+(1-p_{u}^{2})p_{g}^{2}g^{2}\sigma_{d}^{2}+(1-p_{u}^{2}p_{g}^{2})\sigma_{r}^{2}).\nonumber 
\end{align}
Because the expected original pixel value $\mu_{x}$ in the Gaussian
term is impossible to obtain, we approximate it as $\mu_{x}=x_{pre}$.
Based on this equation, we can precisely simulate as if exposure time,
light intensity, and analog gain were $p_{e}$, $p_{i}$, and $p_{g}$
times. Then, let's come back to contrast, brightness, and hue augmentation.
When contrast is multiplied by $p_{c}$ and brightness is changed
by $p_{b}$, it can be expressed as, 
\begin{equation}
x_{new}=p_{c}x_{pre}+p_{b}.\label{eq:12}
\end{equation}
This function is represented as multiplication by $\frac{\left(p_{c}x_{pre}+p_{b}\right)}{x_{pre}}$.
Therefore, noise-accounted contrast and brightness augmentation is
finally defined as, 
\begin{equation}
\begin{cases}
random\:p_{c},\:p_{b}\\
random\:p_{u},\:p_{g}\:(where\:p_{u}p_{g}=\frac{\left(p_{c}x_{pre}+p_{b}\right)}{x_{pre}},\:p_{u},p_{g}>0)\\
Eq.(\ref{eq:11})\:(\mu_{x}\leftarrow x_{pre})
\end{cases}.\label{eq:contrast}
\end{equation}
 We can also convert hue by changing $p_{c}$ and $p_{b}$ per color
filter position in the RAW bayer.

\subsubsection{Blur Augmentation}

Next, we introduce noise-accounted blur augmentation. Usual blur augmentation
makes noise smaller than actual blur because the noise $n_{d}$ and
$n_{r}$ are smoothed out although their noise amounts, in reality,
are not related to how fast you shake a camera or how fast objects
move. Only the photon number-related noise is smoothed out in actual
motion blur. The actual blurred pixel is expressed as, 
\begin{equation}
x_{new}\sim\mathcal{N}\left(g\alpha\sum_{k}w_{k}\bar{u_{k}},\:g^{2}\alpha^{2}\sum_{k}w_{k}\bar{u_{k}}+g^{2}\sigma_{d}^{2}+\sigma_{r}^{2}\right),\label{eq:14}
\end{equation}
 where $\sum_{k}w_{k}=1$ is the blur kernel. With similar equation
manipulation from Eq. (\ref{eq:9}) to Eq. (\ref{eq:11}) , noise-accounted
blur augmentation is 
\begin{align}
x_{new} & \sim\mathcal{N}\left(\sum_{k}w_{k}g\alpha\bar{u_{k}},\:\sum_{k}w_{k}^{2}(g^{2}\alpha^{2}\bar{u_{k}}+g^{2}\sigma_{d}^{2}+\sigma_{r}^{2})\right)\nonumber \\
 & \:\;\:\:+\mathcal{N}(0,\:-\sum_{k}w_{k}^{2}(g^{2}\alpha^{2}\bar{u_{k}}+g^{2}\sigma_{d}^{2}+\sigma_{r}^{2})\nonumber \\
 & \:\;\:\:\;\:\:\:\:\:\:\;\:\:\;\:\:\:\:+g^{2}\alpha^{2}\sum_{k}w_{k}\bar{u_{k}}+g^{2}\sigma_{d}^{2}+\sigma_{r}^{2})\label{eq:blur}\\
 & =\sum_{k}w_{k}x_{pre}+\mathcal{N}(0,\:g\alpha\sum_{k}(1-w_{k})w_{k}x_{pre,k}\nonumber \\
 & \:\;\:\;\:\:\;\:\:\:\:\:\:\;\:\:\;\:\:\:\:\:\:\;\:\:\;\:\:\;\:\:\:\;\:\:\;\:\:+(1-\sum_{k}w_{k}^{2})(g^{2}\sigma_{d}^{2}+\sigma_{r}^{2})).\nonumber 
\end{align}
We account for prior noise but not for prior blur amounts because
most images in usual datasets are not very blurry. Furthermore, it
is difficult to estimate the prior blur amount.

In addition, please note that augmentation to make images clean (make
them brighter and deblurred) is inevitably difficult with noise-accounted
RAW augmentation. Clipping noise variance in Eq. (\ref{eq:11}) to
zero forcibly enables brightening, but brightening too much causes
a mismatch of noise domain.

\section{Evaluation}

\subsection{Dataset}

\begin{figure*}
\centering

\def\svgwidth{1.9\columnwidth}

\scriptsize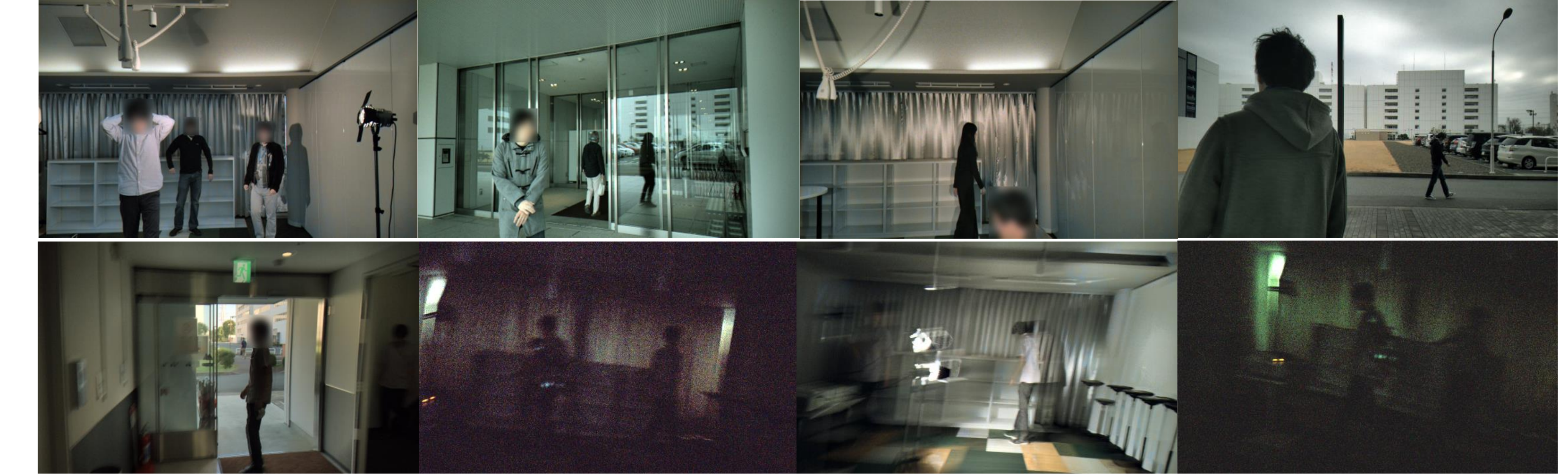

\caption{Examples of the introduced dataset. The upper row shows the training
dataset collected in simple environments while the lower row shows
the challenging test dataset collected in various environments from
dark to bright, HDR, and with or without handshake blur.}

\label{fig:dataset}
\end{figure*}

Although our method can be applied to any computer vision task, we
choose a human detection task as a target because of its wide usage.
We prepared a RAW image dataset for human detection task captured
with an internally developed sensor. As mentioned earlier, our objective
is to train image recognition models that work in various environments
only using a training dataset in simple environments. So, most training
images are taken under normal light conditions with fixed camera positions
in several environments. Note that moderately dark and HDR images
are also included to some extent in the training dataset. The analog
gain is set to 6dB outdoors, 12dB indoors, and 32dB at night in moderate
darkness to generate realistic easy images without auto-exposure.
Test images, on the other hand, are taken in HDR or extremely dark
environments. In addition, about 50\% of them are captured with a
strong camera shake. Moreover, the analog gain is chosen from 3dB,
6dB, 12dB, and 24dB regardless of the environment. Both datasets are
captured with around 1 fps to increase diversity among images. 

We manually annotate the human bounding boxes of both training and
test data. Because precise annotation of test data on sRGB is impossible
due to the noise and blur, we apply an offline ISP to each image and
then annotate the bounding boxes. We manually set adequate ISP parameters
per image and had to change the parameters several times to grasp
the entire image. To avoid annotating large training datasets in this
time-consuming way, it is desirable to train the model with a simple
dataset. In total, 18,880 images are collected for training and 2,800
for testing. The examples are shown in Fig. \ref{fig:dataset}.

\subsection{Implementation Details}

We mainly test with TTFNet \cite{liu2020training} whose backbone
is ResNet18 \cite{he2016deep}. The network is trained for 48 epochs
from scratch using Adam optimizer \cite{kingma2014adam} and a cosine
decay learning rate scheduler with a linear warmup \cite{loshchilov2016sgdr}
for the first 1,000 iterations whose maximum and minimum learning
rates are 1e-3 and 1e-4. As to an ISP, a simple software ISP consisting
of only a gamma tone mapping is implemented. In detail, two types
of gamma tone mapping are implemented. One is the simplest gamma tone
mapping, $y=x^{\frac{1}{\gamma}}\:(0\leq x\leq1)$. The $\gamma$
is set to 5 after tuning with a rough grid search manner. The other
is a gamma tone mapping parameterized with three parameters \cite{mosleh2020hardware}.
Because the grid search for three parameters is time-consuming, we
tuned the parameters with backpropagation together with the detector's
weights as performed in \cite{onzon2021neural,wu2019visionisp}. Other
ISP functions are not used because they are known to have less impact
on image recognition compared with tone mapping \cite{hansen2021isp4ml}.
We also prepare an elaborated black-box ISP consisting of many functions
in addition to a tone mapping function. The parameters are tuned for
human perceptual quality by experts. We only use the elaborated black-box
ISP under the conventional training pipeline. In other words, experiments
are performed under ISP-augmentation-detection order with the elaborated
black-box ISP due to the hardware limitation. If contrast augmentation
is used, the hue is also changed with a probability of 50\%. In detail,
the contrast factor per color channel $p_{c,h=\{R,G,B\}}$ is randomized
from the base contrast factor $p_{c,base}$ by (0.8$p_{c,base}$,
1.2$p_{c,base}$) after $p_{c,base}$ is randomly decided. If blur
augmentation is used, a random-sized blur kernel with a probability
of 50\% is used. Random shift and random scale augmentation whose
maximum transformations are 10\% and 3\% of the input size are also
applied with a probability of 80\% before color jitter augmentation.
The input size to the detector is $\left(576,\,352,\,3\right)$. The
performance of the detector is evaluated with average precision (AP@0.5:0.95)
\cite{lin2014microsoft}.

\subsection{Calibration of the Noise Model\label{subsec:Calibration-of-the}}

\begin{figure}
\centering

\def\svgwidth{1.0\columnwidth}

\scriptsize
\begingroup%
  \makeatletter%
  \providecommand\color[2][]{%
    \errmessage{(Inkscape) Color is used for the text in Inkscape, but the package 'color.sty' is not loaded}%
    \renewcommand\color[2][]{}%
  }%
  \providecommand\transparent[1]{%
    \errmessage{(Inkscape) Transparency is used (non-zero) for the text in Inkscape, but the package 'transparent.sty' is not loaded}%
    \renewcommand\transparent[1]{}%
  }%
  \providecommand\rotatebox[2]{#2}%
  \newcommand*\fsize{\dimexpr\f@size pt\relax}%
  \newcommand*\lineheight[1]{\fontsize{\fsize}{#1\fsize}\selectfont}%
  \ifx\svgwidth\undefined%
    \setlength{\unitlength}{755.23178101bp}%
    \ifx\svgscale\undefined%
      \relax%
    \else%
      \setlength{\unitlength}{\unitlength * \real{\svgscale}}%
    \fi%
  \else%
    \setlength{\unitlength}{\svgwidth}%
  \fi%
  \global\let\svgwidth\undefined%
  \global\let\svgscale\undefined%
  \makeatother%
  \begin{picture}(1,0.3701863)%
    \lineheight{1}%
    \setlength\tabcolsep{0pt}%
    \put(0,0){\includegraphics[width=\unitlength,page=1]{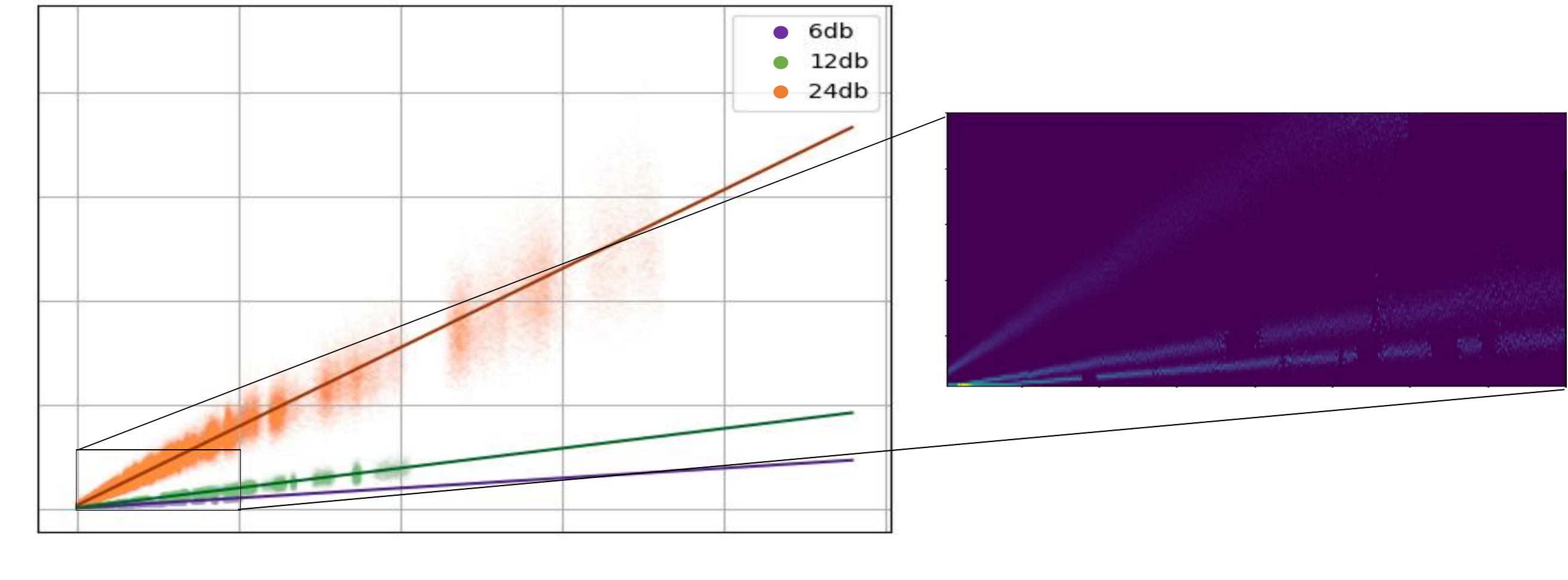}}%
    \put(0.27637833,0.00362782){\color[rgb]{0,0,0}\makebox(0,0)[t]{\lineheight{1.25}\smash{\begin{tabular}[t]{c}mean (expected) pixel value\end{tabular}}}}%
    \put(0.00184483,0.2031389){\color[rgb]{0,0,0}\rotatebox{-90}{\makebox(0,0)[t]{\lineheight{1.25}\smash{\begin{tabular}[t]{c}noise variance\end{tabular}}}}}%
  \end{picture}%
\endgroup%

\caption{Sensor noise model calibration result. The dots in the left graph
are the mean and variance pairs for each pixel, and the lines are
the estimated linear relationship for each analog gain. Since large
numbers of dots are plotted, they seem to be widely spread. On the
other hand, the histogram plot per analog gain (right) shows a clear
difference between them.}

\label{fig:calib_result}
\end{figure}

For each analog gain of 6dB, 12dB, and 24dB, two burst sequences are
captured with different illumination. Each sequence consists of 100
images. $24\times24$ Bayer pixels are sampled from each of the 24
color patches to calculate the mean and variance. In total $2\times24\times24\times24$
pairs of mean and variance sets are obtained per analog gain to estimate
the noise model. Thanks to the various color filters, exposure values,
and color patches, two sequences are enough to ensure diversity. The
lines in Fig. \ref{fig:calib_result} show the estimated linear relationship
in Eq. (\ref{eq:6}). The coefficients of determination, $R^{2}$,
for these line estimations are 0.9833, 0.9884, 0.9862 for 6, 12, and
24dB respectively. High values of $R^{2}$ indicate that the noise
intensity is well modeled with respect to illumination intensity.
Also, $R^{2}$ of the Eq. (\ref{eq:7-1}) and Eq. (\ref{eq:8}) are
$1.0000$ and $0.9984$. This means the noise intensity is well modeled
with respect to the analog gain. Based on the above, our noise model
and the calibration method are well suited to the sensor in terms
of noise intensity.

\begin{figure}
\centering

\def\svgwidth{1.0\columnwidth}

\scriptsize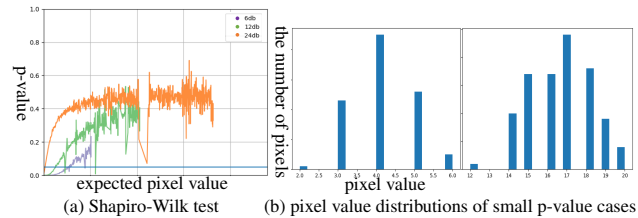

\caption{The result of the Shapiro-Wilk \cite{shapiro1972approximate} test
for each expected pixel value. When testing whether the 100 pixel
values at each location correspond to a Gaussian distribution, most
p-values are larger than 0.05 (a). The right (b) shows the cases where
p \textless{} 0.05. It indicates the small p-values come from sparsity,
not skew.}

\label{fig:shapiro}
\end{figure}

Next, we check the validity of the shape of the distribution. All
noises were assumed to follow a Gaussian distribution. Especially,
it is unclear whether the approximation $\mathcal{P}\left(\bar{u}\right)\fallingdotseq\mathcal{N}\left(\bar{u},\bar{u}\right)$
\cite{foi2008practical} is true. Therefore, the Shapiro-Wilk test
\cite{shapiro1972approximate} is performed. If the p-value of the
test is higher than 0.05, it indicates that we cannot reject the null
hypothesis that the data are normally distributed with more than a
95\% confidence interval. Fig. \ref{fig:shapiro}(a) shows that most
of them are higher than 0.05, but some results for dark pixels are
less than 0.05. However, the distributions of the dark pixels are
like Fig. \ref{fig:shapiro}(b). It is not very skewed and the sparsity
causes the small p-value. Therefore, we conclude that all the noise
sources can be regarded as Gaussian noise.

Based on the above, our sensor noise model and the calibration method
represent the sensor noise well in terms of both intensity and distribution.

\subsection{Statistical Validation of the Noise Alignment\label{subsec:Statistical-Validation-of}}

Before confirming the effectiveness of the proposed noise-accounted
RAW augmentation to computer vision applications, the statistical
validity is evaluated one more time by utilizing the series of color
checker images. The evaluation method is as follows. First, the contrast
of the sequential images is changed with or without noise consideration.
Second, mean and variance pairs along the sequential dimension are
computed. Third, the distributions of real and converted pairs are
examined. If the real and converted pairs matched well, it means the
converted images have the same noise amount as the original real images. 

\begin{figure}
\centering

\def\svgwidth{1.0\columnwidth}

\scriptsize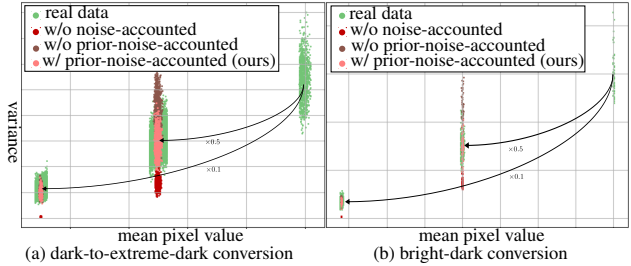

\caption{The statistical validation of the noise alignment. The green dots
represent real image data, and the others are converted from it. The
conversions are \texttimes 0.1 and \texttimes 0.5 contrast conversion
with several methods. The (a) is a dark-to-extreme-dark conversion,
and the (b) is a bright-to-dark conversion.}

\label{fig:statistical_val}
\end{figure}

Three different noise alignment methods are compared, i.e., no noise
consideration, the usual noise-accounted method that disregards prior
noise in the input \cite{Hong2021Crafting,cui2021multitask,wang2020practical,wei2020physics},
and our proposed noise alignment method. Fig. \ref{fig:statistical_val}
shows the comparison result. It indicates that prior noise consideration
is unnecessary if pixels are darkened considerably. However, a small
contrast factor caused a noise domain gap even if the input is bright
like Fig. \ref{fig:statistical_val} (right). In contrast, our noise-accounted
conversion always converted images with realistic noise. It implies
that our proposed method is suited for various strengths of augmentation.
If prior noise is not accounted for, the inputs always have to be
darkened considerably and already dark images are difficult to use.

\subsection{Augmentation Parameters Tuning}

The optimal augmentation parameters should be different between augmentation
before and after ISP. To make a fair comparison, we roughly tune both
of the augmentation parameters. The strategy is as follows. First,
we search the appropriate range of contrast factor $p_{c}$ and brightness
perturbation $p_{b}$ successively. To be robust to any illumination
of inputs, we re-parameterize $p_{b}$ as $p_{b}=\hat{p_{b}}min(x)$
and randomize the $\hat{p_{b}}$ instead of $p_{b}$. Lastly, we search
for the appropriate max blur distance $p_{d}$. In these experiments,
we do not account for sensor noise. We use the elaborated black-box
ISP for the augmentation after ISP setting and the simplest gamma
function for the augmentation before ISP setting.

The results are shown in Table \ref{tab:tuning}. The best parameter
settings are used in the next section.

\begin{table*}
\caption{The augmentation hyperparameter tuning for a fair comparison between
before and after ISP augmentation. The range of contrast, brightness,
and blur distance are tuned one by one, taking over the previous best
parameters.}

\centering\scalebox{0.828}{

\begin{tabular}{cc|ccccc|ccccc}
\hline 
 &  & \multicolumn{5}{c|}{augmentation after ISP (tuned for the black-box ISP)} & \multicolumn{5}{c}{augmentation before ISP (tuned for the simplest ISP)}\tabularnewline
\hline 
\hline 
\multirow{2}{*}{contrast} & range & 0.2-1 & 0.1-1 & 0.2-5 & 0.1-10 & 0.05-20 & 0.1-1 & 0.02-1 & 0.01-1 & 0.005-1 & 0.01-1.1\tabularnewline
 & AP@0.5:0.95 {[}\%{]} & 41.9 & 41.8 & 43.2 & \textbf{45.1} & 44.5 & 36.9 & 39.8 & \textbf{40.4} & 35.8 & 38.9\tabularnewline
\hline 
\multirow{2}{*}{brightness} & range & 0-0 & -0.1-0.1 & -0.2-0.2 & -0.5-0.5 & -0.7-0.7 & 0-0 & -0.1-0.1 & -0.2-0.2 & -0.5-0.5 & -0.7-0.7\tabularnewline
 & AP@0.5:0.95 {[}\%{]} & 45.1 & 44.5 & 44.5 & \textbf{45.2} & 44.7 & 40.4 & \textbf{40.9} & 39.9 & 39.4 & 39.4\tabularnewline
\hline 
\multirow{2}{*}{blur distance} & range & 0-0 & 0-3 & 0-5 & 0-9 & 0-13 & 0-0 & 0-3 & 0-5 & 0-9 & 0-13\tabularnewline
 & AP@0.5:0.95 {[}\%{]} & 45.2 & \textbf{46.8} & 45.7 & 45.9 & 37.9 & 40.9 & 39.8 & 39.6 & 40.6 & \textbf{43.3}\tabularnewline
\hline 
\end{tabular}}

\label{tab:tuning}
\end{table*}

\subsection{Evaluation of the Noise-Accounted RAW Augmentation}

In this section, the proposed noise-accounted RAW augmentation is
evaluated on training image recognition models. First, as Table \ref{tab:result}
shows, augmentation before ISP drastically improves the accuracy with
the simplest ISP. It suggests the realistic pixel intensity distribution
achieved by augmentation before ISP is the key to improving the accuracy
in wide environments. In the color jitter augmentation only setting,
the accuracy is improved from the general noise alignment method \cite{foi2008practical,makitalo2012optimal,Hong2021Crafting,cui2021multitask,liu2014practical,wei2020physics,punnappurath2022day,zamir2020cycleisp,brooks2019unprocessing}
by considering prior noise. In the color jitter and blur augmentation
setting, noise-accounted color jitter augmentation plus normal blur
augmentation does not improve much from no noise-accounted settings.
Instead, noise alignment in both color jitter and blur augmentation
improves the accuracy. It indicates that random noise is not effective
and realistic noise is important. Comparing the accuracy under the
same simplest gamma tone mapping setting, our proposed noise-accounted
RAW augmentation doubles the accuracy of conventional augmentation
after ISP. Furthermore, when parameterized gamma tone mapping is used
as our simple ISP, the accuracy is even superior to the elaborated
black-box ISP consisting of many functions in addition to a tone mapping
function. As the visualization results in the supplementary material
show, the elaborated black-box ISP outputs more perceivable images.
It suggests that minimizing the domain gap caused by augmentation
is more important than the superiority of the ISP. We might improve
the accuracy more by an elaborated ISP and the proposed augmentation.

\begin{table}
\caption{Evaluation of the noise-accounted RAW augmentation. The color augmentation
contains default hue augmentation plus tuned contrast and brightness
augmentation. The \emph{w/o prior} means the prior input noise is
disregarded like many of the previous noise-accounted image conversion
methods \cite{foi2008practical,makitalo2012optimal,Hong2021Crafting,cui2021multitask,liu2014practical,wei2020physics,punnappurath2022day,zamir2020cycleisp,brooks2019unprocessing}.
Because we adopt the well-established heteroscedastic Gaussian model
Eq. (\ref{eq:2}), it is identical to the noise alignment of \cite{brooks2019unprocessing,punnappurath2022day,foi2008practical,zamir2020cycleisp}.
In this experiment, we also use the parameterized gamma tone mapping
as the simple ISP, although it can't be used in the augmentation after
ISP settings because the gradient from the detection loss is needed
to tune.}

\centering

\scalebox{0.75}{

\begin{tabular}{ccc|c|c|c}
\hline 
\multicolumn{2}{c}{} &  & \multicolumn{3}{c}{mAP@0.5:0.95 {[}\%{]}}\tabularnewline
 &  &  & black-box & \multicolumn{2}{c}{simple ISP }\tabularnewline
\multicolumn{2}{c}{augmentation} & noise & ISP & simplest & parameterized\tabularnewline
\hline 
\hline 
\multirow{4}{*}{Color} & after & - & 45.2 & 19.3 & -\tabularnewline
\cline{2-6} \cline{3-6} \cline{4-6} \cline{5-6} \cline{6-6} 
 & \multirow{3}{*}{%
\begin{tabular}{c}
before \tabularnewline
(ours)\tabularnewline
\end{tabular}} & - & - & 40.9 & 44.4\tabularnewline
 &  & w/o prior  & - & 43.5 & 47.7\tabularnewline
 &  & ours & - & \textbf{44.6} & \textbf{48.1}\tabularnewline
\hline 
\multirow{4}{*}{%
\begin{tabular}{c}
Color\tabularnewline
+\tabularnewline
Blur\tabularnewline
\end{tabular}} & after & - & 46.8 & 20.4 & -\tabularnewline
\cline{2-6} \cline{3-6} \cline{4-6} \cline{5-6} \cline{6-6} 
 & \multirow{3}{*}{%
\begin{tabular}{c}
before\tabularnewline
(ours)\tabularnewline
\end{tabular}} & - & - & 43.3 & 43.8\tabularnewline
 &  & ours$\dagger$ & - & 43.4 & 47.9\tabularnewline
 &  & ours & - & \textbf{45.3} & \textbf{48.3}\tabularnewline
\hline 
\end{tabular}

}

\scalebox{0.75}{

$\dagger$: The noise alignment is only applied to the color jitter
augmentation.

}

\label{tab:result}
\end{table}

As mentioned earlier, there are noise dealing works in noise-related
fields like denoising. We compare ours with these methods in the detection
task. One is the K-Sigma transform \cite{wang2020practical}, a kind
of noise domain generalization. It normalizes images so that the pixel
value and the standard deviation of the noise are linearly correlated.
The other is noise amount notification with a concatenation of the
noise variance map \cite{brooks2019unprocessing}. To follow the previous
setup, direct RAW input without an ISP is also compared. Color jitter
and blur augmentation are also applied to the methods different from
the previous papers for a fair comparison. Table \ref{tab:result-1}
shows the comparison results. As to the K-Sigma transform, simply
applying \textquotedbl aug.\textquotedbl{} before or after the K-Sigma
transforms gives a better result. However, there is a theoretical
problem in both cases. If \textquotedblleft aug.\textquotedblright{}
is applied after the transform, the linear relation between pixel
value and noise amount is retained but pixel intensity becomes inconsistent.
On the other hand, applying \textquotedblleft aug.\textquotedblright{}
before the transform makes the noise amount unrealistic. Changing
the augmentation to \textquotedblleft our aug.\textquotedblright{}
makes the intensity and noise realistic and improves the accuracy.
From the experiment, we find the proposed augmentation boosts previous
noise-dealing methods if ISPs are used. However, if ISPs are not used,
noise accounted augmentation slightly deteriorated the accuracy. We
argue that it is because the intensity distributions of RAW images
are too difficult and unrealistic clean images might help training.
However, the overall accuracy is lower than with ISPs. Also, for the
detection task, only our proposed method is enough. It might be because,
unlike the denoising task which should focus on noise, it is important
to make the detector focus on pixel intensity distribution for the
detection task.

\begin{table}
\caption{The comparison results with other noise-dealing techniques. The \textquotedblleft aug.\textquotedblright{}
means contrast and blur augmentation without noise-accounted and \textquotedblleft our
aug.\textquotedblright{} means with noise-accounted. We use the simplest
gamma function in the ISP.}

\centering

\begin{tabular}{c|cc}
\hline 
 & \multicolumn{2}{c}{AP@0.5:0.95 {[}\%{]}}\tabularnewline
method & w/o ISP & w/ ISP\tabularnewline
\hline 
\hline 
concat \cite{brooks2019unprocessing} & 16.5 & 21.5\tabularnewline
aug. + concat \cite{brooks2019unprocessing} & \textbf{35.0} & 31.6\tabularnewline
our aug. + concat \cite{brooks2019unprocessing} & 33.7 & 40.4\tabularnewline
K-Sigma \cite{wang2020practical} & 14.3 & 27.5\tabularnewline
K-Sigma \cite{wang2020practical} + aug. & 25.0 & 34.1\tabularnewline
aug. + K-Sigma \cite{wang2020practical} & 26.6 & 42.1\tabularnewline
our aug. + K-Sigma \cite{wang2020practical} & 26.3 & 44.0\tabularnewline
our aug. & 32.8 & \textbf{45.3}\tabularnewline
\hline 
\end{tabular}

\label{tab:result-1}
\end{table}

\section{Conclusion}

We propose a noise-accounted RAW augmentation method in which augmentation
is applied before ISP to minimize the luminance domain gap and a sensor
noise model is taken into account to minimize the noise domain gap.
Unlike previous noise-accounted methods, ours takes the prior input
noise into account. It minimizes the domain gap more and enables the
use of already noisy images as training data. Thanks to the realistic
augmentation, our method improves the detection accuracy in difficult
scenes compared to the conventional methods. In the future, we would
like to investigate whether the proposed augmentation with an elaborate
ISP improves computer vision performance even further. Also, we would
like to check the effectiveness against other image recognition tasks,
such as classification or segmentation, by preparing datasets. We
believe it is effective because our method is task-independent. Lastly,
we are glad if this work sheds light on the importance of RAW images.

\url{}

{\small
\bibliographystyle{ieee_fullname}
\bibliography{Rawgment_arxiv_ver9.bbl} 
}{\small\par}

\appendix

\part*{Appendices}

\section{Visualization Results}

\begin{figure*}
\centering

\setlength{\tabcolsep}{0.5pt}


\begin{tabular}{ccccccc}
 & \includegraphics[width=0.142\textwidth]{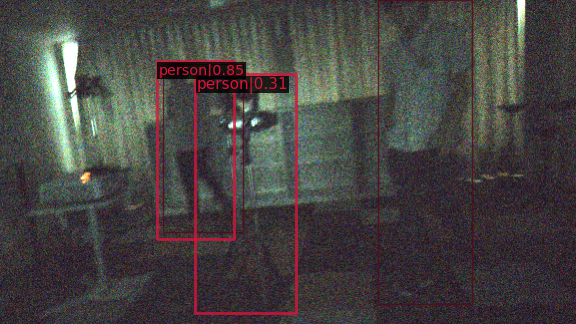} & \includegraphics[width=0.142\textwidth]{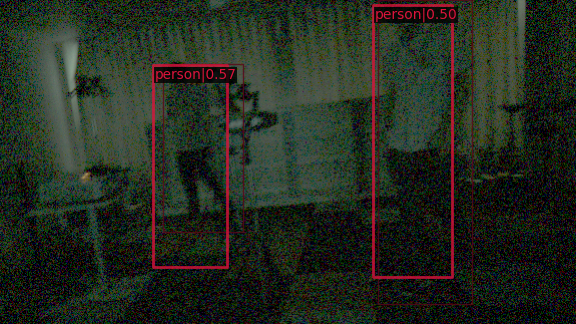} & \includegraphics[width=0.142\textwidth]{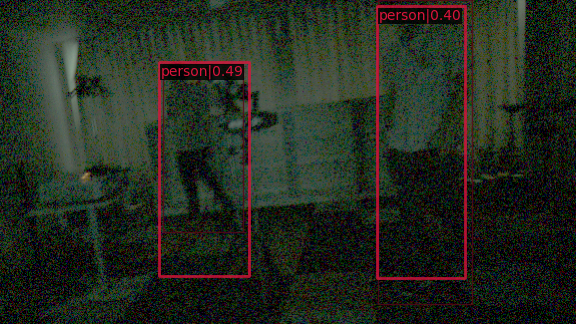} & \includegraphics[width=0.142\textwidth]{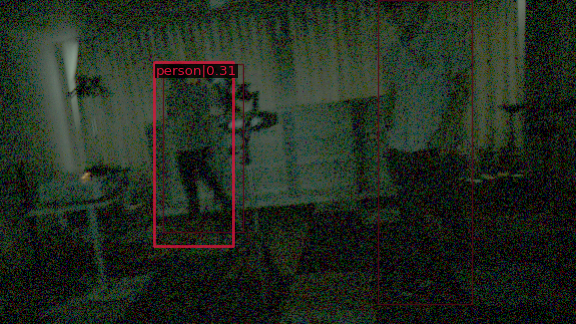} & \includegraphics[width=0.142\textwidth]{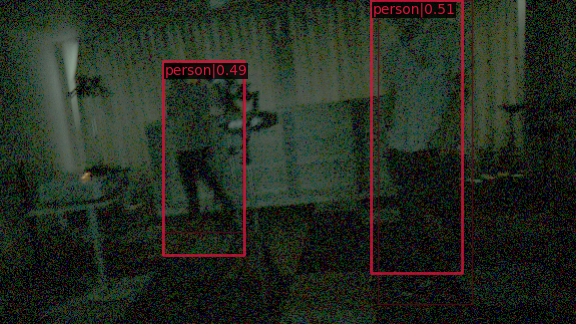} & \includegraphics[width=0.142\textwidth]{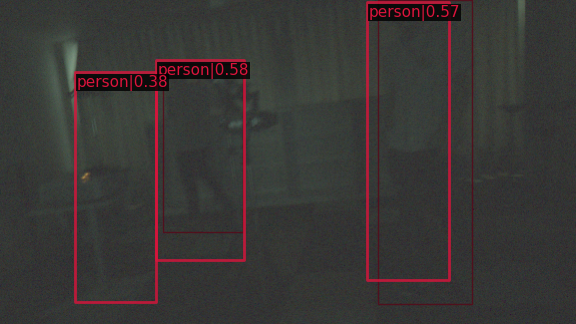}\tabularnewline
 & \includegraphics[width=0.142\textwidth]{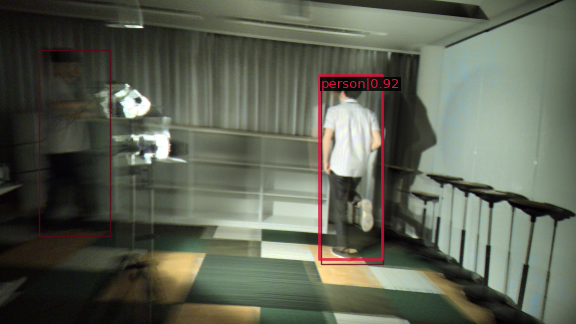} & \includegraphics[width=0.142\textwidth]{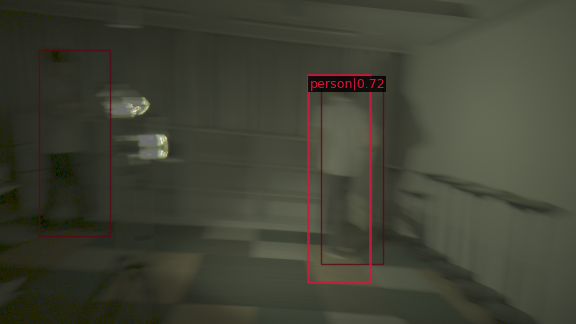} & \includegraphics[width=0.142\textwidth]{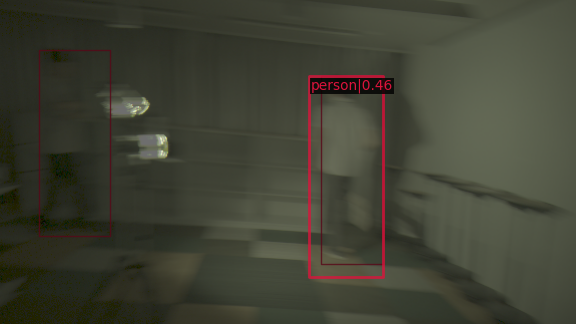} & \includegraphics[width=0.142\textwidth]{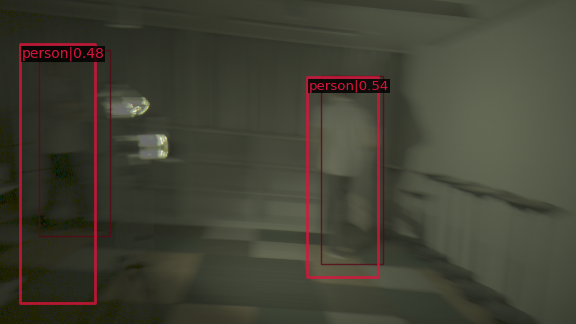} & \includegraphics[width=0.142\textwidth]{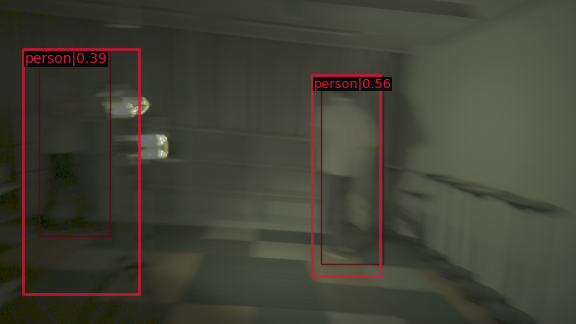} & \includegraphics[width=0.142\textwidth]{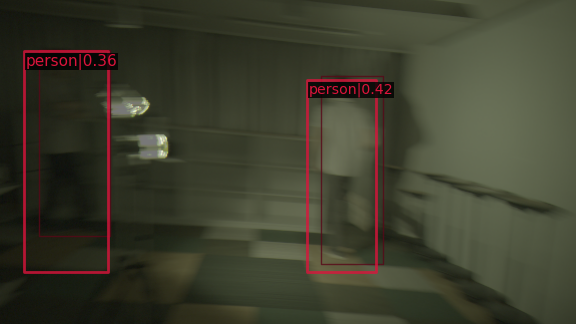}\tabularnewline
 & \includegraphics[width=0.142\textwidth]{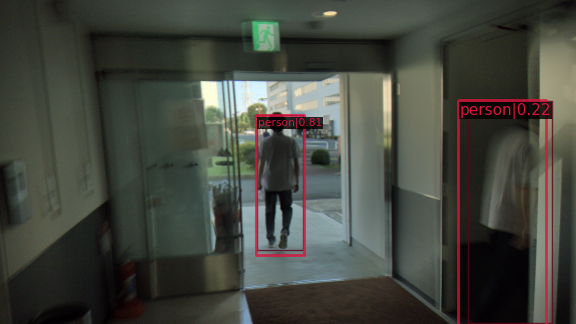} & \includegraphics[width=0.142\textwidth]{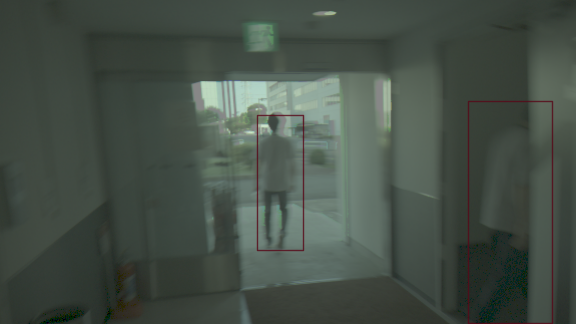} & \includegraphics[width=0.142\textwidth]{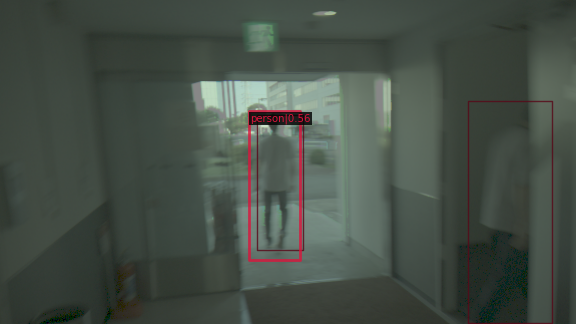} & \includegraphics[width=0.142\textwidth]{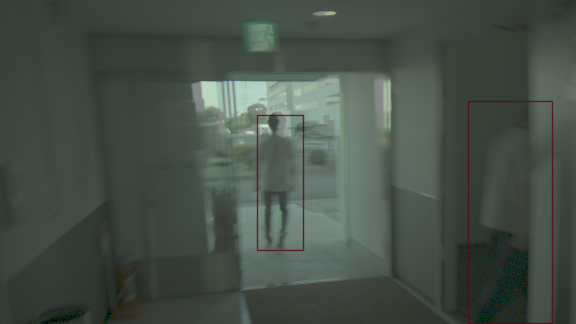} & \includegraphics[width=0.142\textwidth]{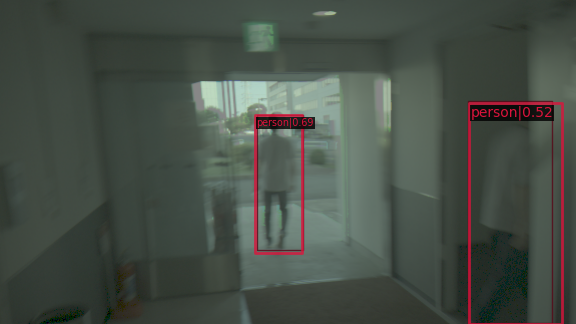} & \includegraphics[width=0.142\textwidth]{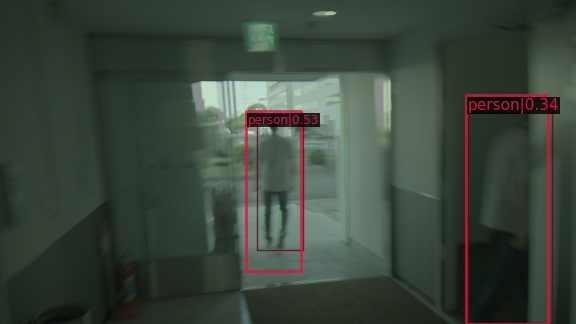}\tabularnewline
 & \includegraphics[width=0.142\textwidth]{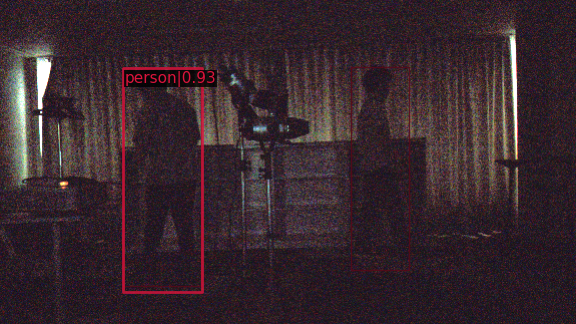} & \includegraphics[width=0.142\textwidth]{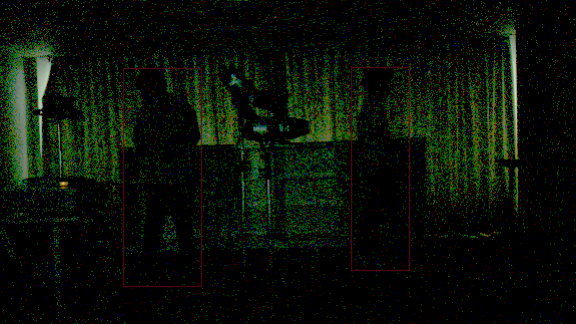} & \includegraphics[width=0.142\textwidth]{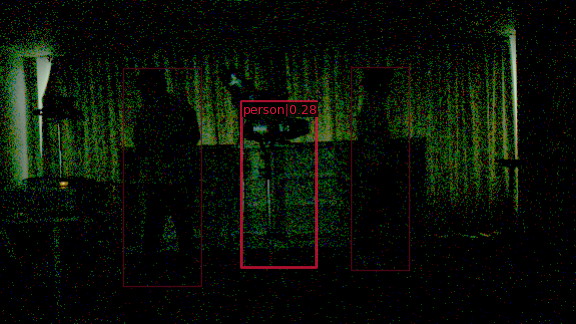} & \includegraphics[width=0.142\textwidth]{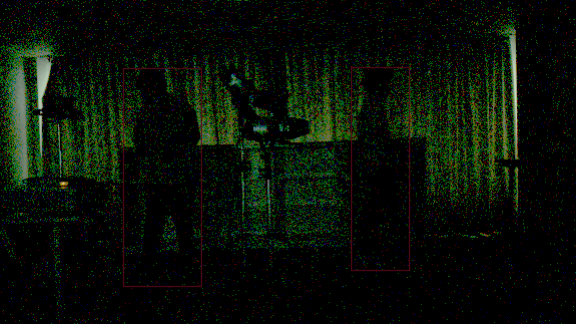} & \includegraphics[width=0.142\textwidth]{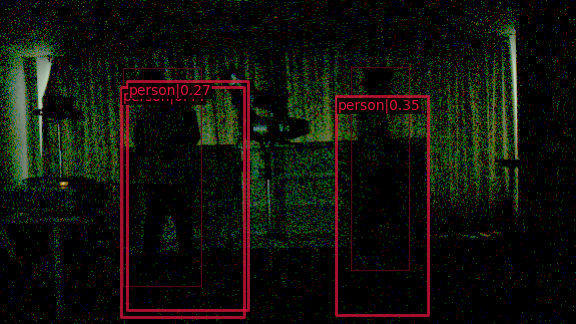} & \includegraphics[width=0.142\textwidth]{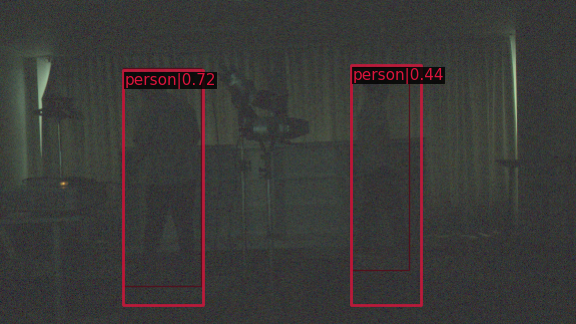}\tabularnewline
\scalebox{0.8}{%
\begin{tabular}{c}
ISP\tabularnewline
augmentation order\tabularnewline
noise-accounted\tabularnewline
\end{tabular}} & \scalebox{0.8}{%
\begin{tabular}{c}
black-box ISP\tabularnewline
after ISP\tabularnewline
w/o\tabularnewline
\end{tabular}} & \scalebox{0.8}{%
\begin{tabular}{c}
simplest gamma\tabularnewline
after ISP\tabularnewline
w/o\tabularnewline
\end{tabular}} & \scalebox{0.8}{%
\begin{tabular}{c}
simplest gamma\tabularnewline
before ISP (ours)\tabularnewline
w/o\tabularnewline
\end{tabular}} & \scalebox{0.8}{%
\begin{tabular}{c}
simplest gamma\tabularnewline
before ISP (ours)\tabularnewline
w/o prior\tabularnewline
\end{tabular}} & \scalebox{0.8}{%
\begin{tabular}{c}
simplest gamma\tabularnewline
before ISP (ours)\tabularnewline
w/ (ours)\tabularnewline
\end{tabular}} & \scalebox{0.8}{%
\begin{tabular}{c}
parameterized gamma\tabularnewline
before ISP (ours)\tabularnewline
w/ (ours)\tabularnewline
\end{tabular}}\tabularnewline
\end{tabular}

\caption{The visualization of the detection results. To make a fair comparison,
we set an adequate confidence threshold per model. Specifically, we
adjuste the threshold to achieve a precision@0.5 value of 80\%. The
darker bounding boxes represent the ground truth, while the brighter
ones represent the prediction result.}

\label{fig:result}
\end{figure*}

In Fig. \ref{fig:result}, the detection results are drawn on the
output of the corresponding ISPs. The proposed method shows a significant
improvement in accuracy under the condition that the simplest gamma
tone mapping is used as an ISP. In addition, the accuracy of the proposed
method is the best despite the use of the simple ISP with limited
visibility against a rich black-box ISP because of the effective noise-accounted
RAW augmentation.

\section{Additional Experiments}

\subsection{Versatility to Different Detectors}

TTFNet \cite{liu2020training} is used as a detector in the main paper
because of the low training cost. In this section, the versatility
of the proposed noise-accounted RAW augmentation is checked. As a
different type of detector, we choose DeformableDETR \cite{zhu2020deformable}
as a detector. Also, we change the backbone to ResNet50 \cite{he2016deep}
to check the proposed method's effectiveness with a larger model.
Furthermore, the backbone is pre-trained with ImageNet \cite{deng2009imagenet}
to compare with the best accuracy. Other experimental setups are the
same as those with TTFNet.

The result is shown in Fig. \ref{tab:result-2}. Because a larger
detector with the pre-trained backbone is used, all methods have improved
accuracy, but there is still a great improvement from the conventional
augmentation after ISP setup to the proposed noise-accounted RAW augmentation
when the simplest ISP is used. Moreover, if parameterized gamma tone
mapping and the proposed augmentation are used, the accuracy is even
improved from the result with the elaborated black-box ISP, which
should benefit most from the pre-training with sRGB images.

The future work is to check the effectiveness of the combination of
the black-box ISP and the proposed augmentation by implementing the
black-box ISP as software that works on a computer.

\begin{table}
\caption{Evaluation with DeformableDETR \cite{zhu2020deformable} whose backbone
is ResNet50 pre-trained with ImageNet.}

\centering

\scalebox{0.8}{

\begin{tabular}{ccc|c|c|c}
\hline 
\multicolumn{2}{c}{} &  & \multicolumn{3}{c}{mAP@0.5:0.95 {[}\%{]}}\tabularnewline
 &  &  & black-box & \multicolumn{2}{c}{simple ISP}\tabularnewline
\multicolumn{2}{c}{augmentation} & noise & ISP & simplest & parameterized\tabularnewline
\hline 
\hline 
\multirow{3}{*}{%
\begin{tabular}{c}
Color\tabularnewline
+\tabularnewline
Blur\tabularnewline
\end{tabular}} & after & - & 51.6 & 40.2 & -\tabularnewline
\cline{2-6} \cline{3-6} \cline{4-6} \cline{5-6} \cline{6-6} 
 & \multirow{2}{*}{%
\begin{tabular}{c}
before\tabularnewline
(ours)\tabularnewline
\end{tabular}} & - & - & 46.8 & 47.5\tabularnewline
 &  & ours & - & \textbf{51.5} & \textbf{52.0}\tabularnewline
\hline 
\end{tabular}

}

\label{tab:result-2}
\end{table}

\section{Acknowledgements}

We would like to thank Aji Widya and Iheb Begalcem for their helpful
comments to this manuscript.
\end{document}